\documentclass[journal]{IEEEtran}
\usepackage{ifpdf}
\usepackage{graphicx}
\usepackage{multirow}
\usepackage{amsmath,amssymb} % define this before the line numbering.
\usepackage{color}
\begin{document}

\title{Unsupervised Deep Multi-focus Image Fusion}

\author{Xiang~Yan,~\IEEEmembership{Student Member,~IEEE,}
        Syed~Zulqarnain~Gilani,
        Hanlin~Qin,
        and~Ajmal~Mian % <-this % stops a space
\IEEEcompsocitemizethanks{\IEEEcompsocthanksitem Xiang Yan and Hanlin Qin are  with  School of Physics and Optoelectronic Engineering, Xidian University, Xi’an 710071, ShannXi, China.  \protect\
E-mail: (xyan@stu.xidian.edu.cn, hlqin@mail.xidian.edu.cn).
\IEEEcompsocthanksitem Syed Zulqarnain Gilani and Ajmal Mian are with the Department of Computer Science and Software Engineering, The University of Western Australia, Crawley,  WA, 6009, Australia.\protect\ 
E-mail: ({zulqarnain.gilani,ajmal.mian}@uwa.edu.au).}}% <-this % stops a space

%\markboth{Journal of \LaTeX\ Class Files,~Vol.~14, No.~8, August~2018}%
%\markboth{IEEE TRANSACTIONS ON IMAGE PROCESSING}%
%{Shell \MakeLowercase{\textit{et al.}}: Bare Advanced Demo of IEEEtran.cls for IEEE Computer Society Journals}

\IEEEtitleabstractindextext{%
\begin{abstract}
Convolutional neural networks have recently been used for  multi-focus image fusion. However, due to the lack of labelled data for supervised training of such networks, existing methods have resorted to adding Gaussian blur in focused images to simulate defocus and generate synthetic training data with ground-truth for supervised learning. Moreover, they classify pixels as focused or defocused and leverage the results to construct the fusion weight maps which then necessitates a series of post-processing steps. In this paper, we present unsupervised end-to-end learning for directly predicting the fully focused output image from multi-focus input image pairs. The proposed approach uses a novel CNN architecture trained to perform fusion without the need for ground truth fused images and exploits the image structural similarity (SSIM) to calculate the loss; a metric that is widely accepted for fused image quality evaluation. Consequently, we are able to utilize {\em real} benchmark datasets, instead of simulated ones, to train our network. The model is a feed-forward, fully convolutional neural network that can process images of variable sizes during test time. Extensive evaluations on benchmark datasets show that our method outperforms existing state-of-the-art in terms of visual quality and objective evaluations.
\end{abstract}

\begin{IEEEkeywords}
Multi-focus image fusion, Convolution Neural Network, Unsupervised learning.
\end{IEEEkeywords}}

\maketitle

% is not selected <OR> if conference mode is selected - because compsoc
% conference papers position the abstract like regular (non-compsoc)
\IEEEdisplaynontitleabstractindextext

% \fi
\IEEEpeerreviewmaketitle

\ifCLASSOPTIONcompsoc
\IEEEraisesectionheading{\section{Introduction}\label{sec:introduction}}
\else
\section{Introduction}
\label{sec:introduction}
\fi

%\IEEEPARstart{M}{ulti-focus} Image Fusion (MFIF) aims at reconstructing a fully focused image from two or more partly focused images acquired by imaging devices like a digital single-lens reflex camera. from the same scene since most imaging systems have a limited depth-of-field so that scene content within a limited distance from the imaging plan remains in focus. Specifically, the objects closer or further than that appears as blurred or out-of-focus in the image. 

\IEEEPARstart{M}{ost} imaging systems, for instance digital single-lens reflex cameras, have a limited depth-of-field such that the scene content within a limited distance from the imaging plane remains in focus. Specifically, objects closer to or further away from the point of focus appear as blurred (out-of-focus) in the image.  Multi-Focus Image Fusion (MFIF) aims at reconstructing a fully focused image from two or more partly focused images of the same scene. MFIF techniques have wide ranging applications in the fields of surveillance, medical imaging, computer vision, remote sensing and digital imaging \cite{li2008multifocus,saha2013,gangapure2015,phamila2014,kong2014}. Though interesting and seemingly trivial, multi-focus image fusion is a challenging task~\cite{li2008multifocus}.

%==============
%Example-based MFIF method have demonstrated the good performance by learning a classifier to distinguish the focused pixels from the unfocused pixels using some focused image datasets to simulate unfocused images so that generate training datasets~\cite{li2008multifocus}. Some learning algorithms have been applied to learn such classifier, including dictionary learning~\cite{nejati2015multi}.\par

%--------------------- YOU DO NOT NEED THIS FIGURE
% \begin{figure*}
% \begin{center}
%  \includegraphics[width=0.95\linewidth]{TIP_figure1.jpg}
%  \end{center}
%   \caption{ {\bf Framework of the existing multi-focus image fusion based on CNN}. The pipeline consists of pixel classification from input images, focus measurement, calculating initial decision map, optimizing the  decision maps and finally fusing  the images.}
%  \label{fig:1}
%    \end{figure*}
%%-----------------------------------------
The advent of Convolutional Neural Networks (CNNs) has seen a revolution in Computer Vision in tasks ranging from object recognition~\cite{simonyan2014very,he2016deep}, semantic segmentation~\cite{long2015fully,noh2015learning}, action recognition~\cite{simonyan2014two,wang2016temporal}, optical flow~\cite{dosovitskiy2015flownet,lai2017semi} to image super-resolution~\cite{dong2016image,ledig2016photo,Lai_2017_CVPR}. Recently, Prabhakar et al.~\cite{prabhakar2017deepfuse} used deep learning to fuse multi-exposure image pairs. This was followed by Liu et al.~\cite{liu2017multi} who proposed a Convolutional Neural Network (CNN) as part of their algorithm to fuse multi-focus image pairs. The algorithm learns a classifier to distinguish between ``focused'' and ``unfocused'' images and jointly calculates a fusion weight map. Later, Tang et al.~\cite{tang2017pixel} improved the algorithm by proposing a pixel-CNN (p-CNN) for classification of ``focused'' and ``defocused'' pixels in a pair of multi-focus images. It is well known that the performance of CNNs depends on the availability of large training data with labels~\cite{gilani2018}.  Liu et al.~\cite{liu2017multi} and Tang et al.~\cite{tang2017pixel} addressed this problem by simulating blurred versions of benchmark datasets used for image recognition. Unfocused images were generated by adding Gaussian blur in randomly selected patches making their training dataset unrealistic. Furthermore, since their method is based on calculating weight fusion maps after learning a classifier, it does not provide an end-to-end solution. This necessitates some post-processing steps for improving the results. Finally, in most well known deep networks~\cite{taigman2014,schroff2015} the input image size is restricted to the training image size. For instance DeepFuse~\cite{prabhakar2017deepfuse} creates fusion maps during training and requires the input image size to match the fusion map dimensions. This problem is circumvented by sliding a window over the image and obtaining patches to match the fusion map size. These patches are then averaged to obtain the final weight fusion map of the same size as corresponding source images, thereby introducing redundancy and errors in the final reconstruction.

To address these issues, we present an end-to-end deep network trained on benchmark multi-focus images. The proposed network takes a pair of multi-focus images and outputs the all-focus image. We train our network in an unsupervised fashion precluding the need for a ground truth all focused image. However, this method of training requires a robust loss function. We approach this problem by proposing a multi-focus Structural Similarity (SSIM) quality metric as our loss function.  To the best of our knowledge, this is the first end-to-end unsupervised deep network for  predicting all-focus images from their respective multi-focus image pairs.

In a nutshell our contributions are as follows:

1) {\bf Training dataset.} Instead of using a simulated dataset, we use the benchmark multi-focus image dataset to train our network. Specifically, we use random crops from pairs of multi-focus images thereby generating a large corpus of training data for our network. 

2) {\bf An end-to-end network.} Our proposed network has an end-to-end unsupervised architecture which does not need a reference ground truth image, thus, addressing the issue of lack of ground-truth for training. Furthermore, our architecture differs from existing methods~\cite{liu2017multi} which use deep networks for classification only as part of MFIF.\par

3) {\bf Loss function.} We propose a novel loss function tailored for multi-focus image fusion to train our network. \par

4) {\bf Test images.} Our network can feed test images of any size and  directly output the fused images leading to a more practical value.

5) {\bf Making the network public.} The trained network will be publicly released to encourage replication and verification of our proposed method.

\section{Related work}
Literature is rich in research on image fusion including multi-focus image fusion. Most of the research work can be classified into transform domain based algorithms and spatial domain based algorithm~\cite{li2017pixel}. The spatial domain based algorithms have become popular owing to the advent of CNNs. However, the spatial domain based algorithms compute the weights for each image  either locally or pixel wise. The fused image would then be a weighted sum of the images in the input pair. Here, we present a brief overview of the conventional and CNN based image fusion techniques:\par

\noindent {\bf Transform domain based multi-focus image fusion.} Image fusion has been extensively studied in the past few years. Earlier methods are mostly based on transform domain, owing to their intuitive approach towards this problem. This research mainly focuses on pyramid decomposition~\cite{mitianoudis2007pixel,petrovic2004gradient}, wavelet transform~\cite{hill2002image,lewis2007pixel} and multi-scale geometric analysis~\cite{li2008multifocus,zhang2009multifocus}. Multi-focus image fusion methods mainly include the gradient pyramid (GP)~\cite{petrovic2004gradient}, discrete wavelet transform (DWT)~\cite{pajares2004wavelet}, non-subsampled contourlet transform (NSCT)~\cite{zhang2009multifocus}, shearlet transform (ST)~\cite{miao2011novel},
curvelet transform (CVT)~\cite{guo2012multifocus} among others. Transform domain based multi-focus image fusion method first decomposes the source images into a specific multi-scale domain, then integrates all these corresponding decomposed coefficients to generate a series of comprehensive coefficients. Finally it reconstructs them by performing the corresponding inverse multi-scale transform. For this kind of method, the selection of multi-scale transform approach is significant, at the same time, the fusion rules for high-frequency and low-frequency coefficients also cannot be ignored, since they directly affect the fusion results. In the recent past, Independent Component Analysis (ICA), Principal Component Analysis (PCA), higher-order singular-value decomposition (HOSVD) and sparse representation based methods have also been introduced int he field of multi-focus image fusion. The core idea of these fusion methods is to seek a desirable feature space that can efficiently reflect the activity of image patches. The focus measurement plays a crucial role in these methods.

\noindent{\bf Spatial domain based multi-focus image fusion.} Spatial domain based image fusion algorithms have received significant attention resulting in the development of several image fusion algorithms that operate directly on the source images without converting them into alternative representation. These algorithms apply a fusion rule to the source images to generate an all-in-focus image. Generally, these algorithms can be divided into two groups; pixel based and block (or region) based algorithms~\cite{li2017pixel}. Between the two of them, block or region based multi-focus image fusion methods have been widely adopted, however, they usually select more focused blocks or regions as fused parts. In this way, the focus measure plays a vital role in these fusion methods. Furthermore, this method suffers from the blocking effects in the final fused image. In recent years,the pixel based multi-focus fusion methods have drawn increasing attention form the research community, owing to its capability of extracting details from the source images and preserving the spatial consistency of the fused image~\cite{liu2015multi}. The representative methods include image matting based method~\cite{li2013image}, guided filtering based method~\cite{li-kang2013image} and dense scale-invariant feature transform based methods~\cite{liu2015multi}. These methods have achieved competing results with high computational efficiency.

\noindent{\bf Deep learning for multi-focus image fusion.} More recently, researchers have turned to learning a focus measure without hand crafting using deep CNNs. Generally, neural network based methods divide the source images into patches and feed them into the CNN model along with the focus measure learned for each patch. This method is more robust compared to its conventional counterpart and is without any artifacts since the CNN model is data-driven. Lately, Liu et al.~\cite{liu2017multi} proposed a deep network as a subset of their multi-focus image fusion algorithm. They sourced their training data from popular image classification databases and simply added Gaussian blur to random patches in the image to simulate multi-focus images. The authors used their CNN to classify focused and unfocused pixels and generated an initial focus map from this information. The final all-focus image was generated after post-processing this initial focus map. This step increases the computational cost and makes this method more suitable for parallel GPU processing. Following Liu et al.~\cite{liu2017multi},  Tang et.al~\cite{tang2017pixel} proposed a p-CNN for multi-focus image fusion. The authors leverage the Cifar-10~\cite{Anu:2013} to generate training image sets for  their p-CNN. Specifically, the defocused images are acquired by automatically adding blur to the original images. The output of the model are three probabilities: defocused, focused or unknown for each pixel, which are used to determine the fusion weight map. This step also needs post processing, which is important to obtain a desired fusion weight map.\par

We propose a deep end-to-end neural network model that does not require post processing. Our model is trained on real multi-focus image pairs and  utilizes a no-reference quality metric, multi-focus fusion structural similarity (SSIM), as a loss function to achieve end-to-end unsupervised learning. Our model has three components: feature extraction, fusion and reconstruction and is described in detail in the succeeding paras.\par

%------------------------
\begin{figure*}
\begin{center}
\includegraphics[width=0.95\linewidth]{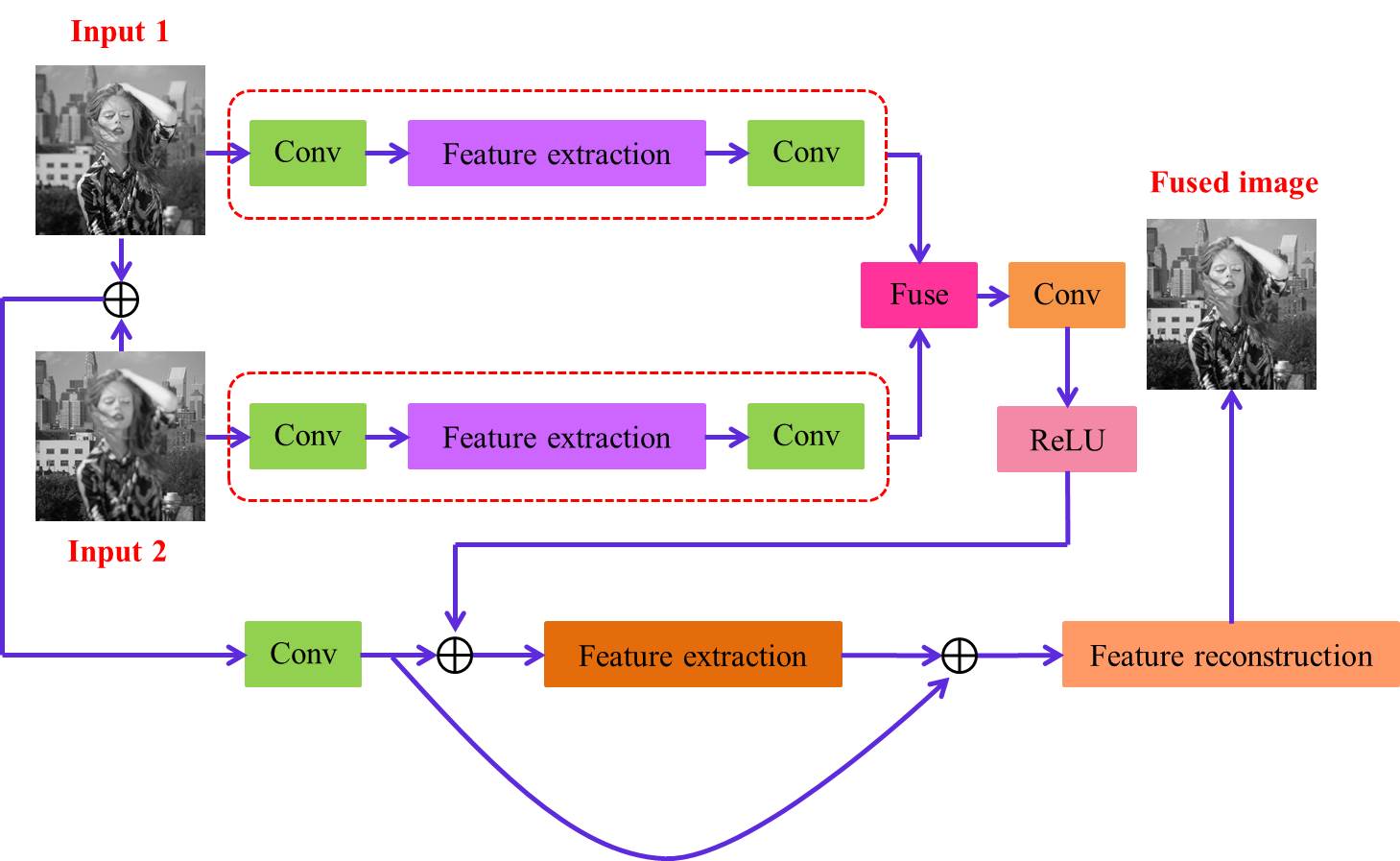}
\end{center}
\caption{ {\bf Detailed network architecture of the proposed multi-focus image fusion network}. Our model consists of three feature extraction networks for extracting non-linear features, a feature reconstruction layer for predicting the fused image, a convolutional layer for feature maps and fused features, and  a transposed convolutional layer for obtaining the same dimensionality as the input image. }
\label{fig:2}
\end{figure*}
%------------------------------
\section{Deep Unsupervised Network For MFIF}

Our main goal is to generate a fused image that is all-in-focus. Given an input multi-focus image pair, our model produces an image that is likely to contain all the pixels in focus. Our method excludes the redundant information contained in the input image pair. In this section, we describe the design of our proposed deep unsupervised Multi-Focus image Fusion Network (\textit{MFNet}).\par

\subsection{Network Architecture}

We propose a deep unsupervised model for the generation of multi-focus image fusion. The network architecture is illustrated in Figure~\ref{fig:2} and comprises of four main sub-networks: three feature extraction sub-network and one feature reconstruction sub-network. 

\subsubsection{Feature Extraction Sub-network}\par
As illustrated in Figure~\ref{fig:2}, each input image from the multi-focus image pair is passed through a feature extraction network (shown in purple ) to obtain high-dimensional non-linear feature maps. However, before passing through this network, the images are convolved with a $3\times3$ kernel and $64$ output channels. The output of the feature extraction network is passed through another convolutional layer without an activation function. The features from these networks for the two images are then fused to obtain a feature map. We also take the average of the two multi-focus image pairs and pass this image through a different feature extraction network (shown in orange in Figure~\ref{fig:2}). The output of this network is then added to the fused output from the fist two feature extraction networks and passed to the feature reconstruction sub-network.

The details of the feature extraction sub-networks are given in Figure~\ref{fig:3}. Each network consists of a stack of multiple convolutional layers followed by rectification layers without any pooling. We use different architectures for the feature extraction sub-networks. The network which takes in the average of the multi-focus images as input has D2 layers and is deeper than the network (having D1 layers) through which the individual images are passed. We have color coded the networks in Figures~\ref{fig:2} and~\ref{fig:3} for ease of cross referencing.

\subsubsection{Feature Reconstruction Sub-network}\par

\begin{figure}[h]
\begin{center}
\includegraphics[width=1.0\linewidth]{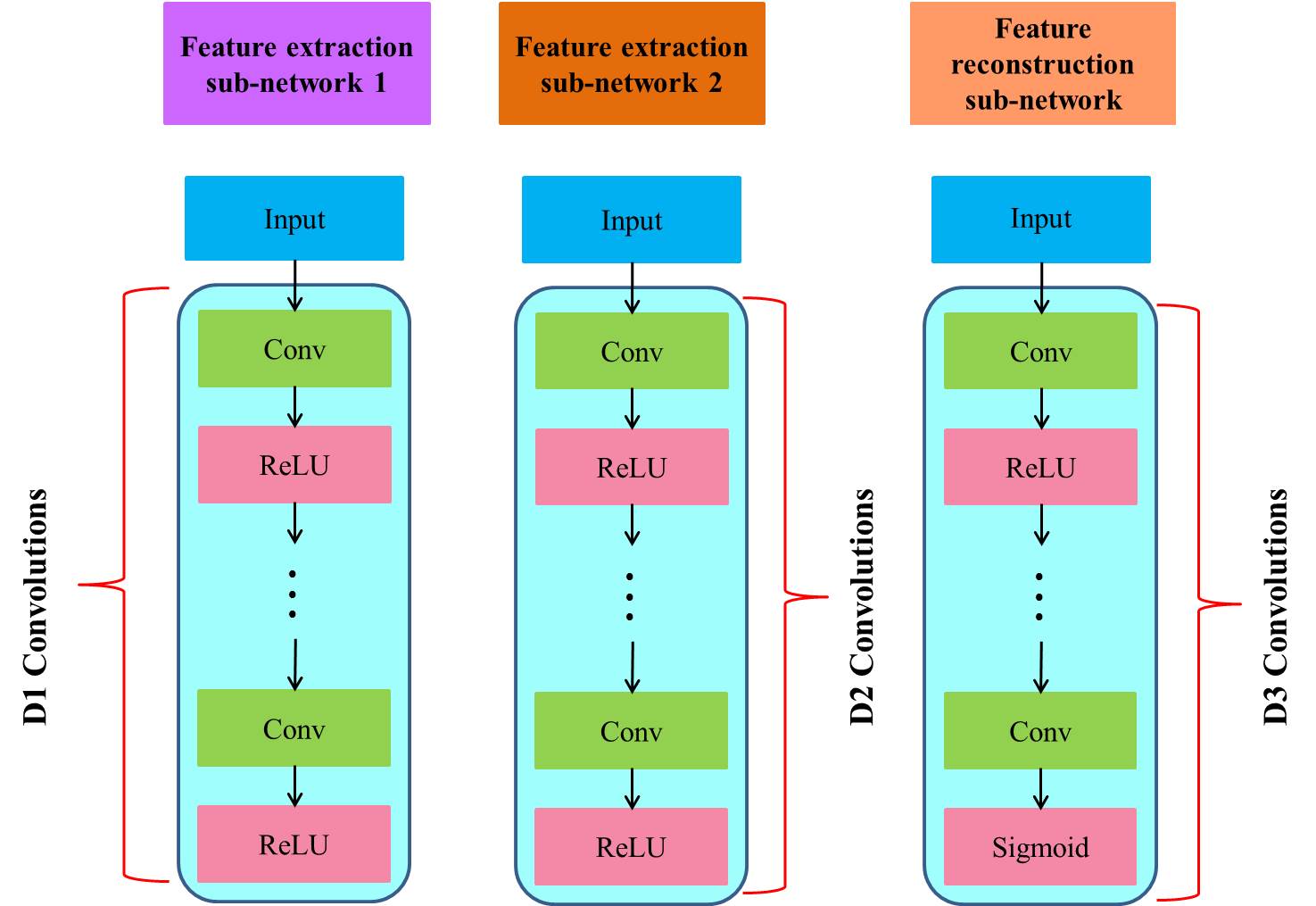}
\end{center}
\caption{ {\bf Structure of our feature extraction and reconstruction sub-networks.} There are D1, D2, D3 convolutional layers in the three networks respectively. The weights of convolutional layers are distinct among these three networks.}
\label{fig:3}
 \end{figure}

The goal of this module is to produce the final fused image. It takes as input the output of the third  feature extraction sub-network and the convolutional features obtained from the two added input images. As illustrated in Figure~\ref{fig:3}, the feature reconstruction network also consists of a cascade of CNNs and is deeper than the feature extraction sub-networks. It comprises seven layers out of which the first six include the leaky rectified linear units (LReLUs) with a negative slope of 0.2 as the non-linear activation functions. The output fusion image is given by the last convolutional layer with sigmoid nonlinearity. Once again, this network is depicted in the same color in Figures~\ref{fig:2} and~\ref{fig:3} for easy cross referencing. 

\subsection{Loss function:}
Our proposed network in trained in an unsupervised fashion in the sense that it does not require ground truth all-in-focus images. Instead, the image structure similarity (SSIM) quality metric is used. The SSIM is often used to evaluate the performance of image fusion algorithms and hence it is natural to use this metric directly as the loss function. Let $ x_{1} $, $ x_{2} $ be the input image pair and $ \theta $ be the set of network parameters to be optimized. Our goal is to learn a mapping function $ g $ for generating an image (fused image) $ \hat{z}=g\left(x_{1},x_{2};\theta\right) $ that is as similar to the desired image (all the pixels in this image are in focus) $ z $ as possible. The network learns the ideal model parameters by optimizing a loss function. We now give details of multi-focus SSIM  and the design of our loss function:

\noindent{\bf (1) } The image structure similarity (SSIM)~\cite{wang2004image} is designed for calculating the structure similarities of different sliding windows in their corresponding positions between two images. Let $ x $ be the reference image and $ y $  be a test image, then the SSIM can be defined as:
 \begin{equation}
 {\rm SSIM}\left ( x,y|w \right )=\frac{\left (2\bar{w}_{x}\bar{w}_{y}+C_{1}  \right )\left (  2\sigma _{w_{x}w_{y}}+C_{2}\right )}{\left ( \bar{w}_{x}^{2}+\bar{w}_{y}^{2}+C_{1} \right )\left ( \sigma _{w_{x}}^{2}+ \sigma _{w_{y}}^{2}+C_{2}\right )} ,
 \end{equation}
where $ C_{1} $ and $ C_{2} $ are two small constants, $ w_{x} $ is a sliding window or the region under consideration in $ x $, $ \bar{w}_{x} $ is the mean of $ w_{x} $,  $ \sigma_{w_{x}}^{2} $ and $ \sigma_{w_{x}w_{y}} $ are the  variance of $ w_{x} $ and covariance of $ w_{x} $ and $ w_{y} $, respectively. The variables $ w_{y} $, $ \bar{w}_{y} $ and $ \sigma_{w_{y}}$ have the same meanings corresponding $ x $.
Note that the value of $ {\rm SSIM}\left(x,y|w\right)\in \left[-1,1\right]$ is used to measure the similarity between $ w_{x} $ and $ w_{y} $. When its value is 1, it means that $ w_{x} $ and $ w_{y} $ are the same.\par
 
\noindent{\bf (2) } Image quality measurement in the local windows. First, we calculate the structure similarities  $ {\rm SSIM}\left(x_{1},\hat{y} |w\right)$ and $ {\rm SSIM}\left(x_{2},\hat{y}|w\right)$ using Equation (1). The constants $ C_{1} $ and $ C_{2} $ are set as $ 1\times 10^{-4} $ and $ 9\times 10^{-4}$, respectively. The size of sliding window is  $ 7 \times 7 $, and it moves pixel by pixel from the top-left to the bottom-right of the image. We use the structural similarity of the input images as matching metric. When the standard deviation std$\left ( x_{1}|w \right )$ of a local window of input $x_{1}$   is equal to larger than the corresponding std$\left ( x_{2}|w \right )$ of input $x_{2}$, it means that the local window image patch of input $ x_{1} $ is more clear. At this time, we can determine the objective function by calculating the image patch similarity. It can be described as follows:\par

\begin{equation}
{\rm Scope } \left(x_{1},x_{2}, \hat{y} |w\right)=\left\{
\begin{array}{lcl}
{\rm SSIM}\left(x_{1},\hat{y}|w\right), \\for\ {{\rm std}\left(x_{1}|w\right)\geq {\rm std}\left(x_{2}|w\right)}\\ 
{\rm SSIM}\left(x_{2},\hat{y}|w\right), \\for\ {{\rm std}\left(x_{1}|w\right)< {\rm std}\left(x_{2}|w\right)}
\end{array} \right.
\end{equation}

\noindent {\bf (3) } Loss function. Based on the value of ${\rm Scope}\left(x_{1},x_{2},\hat{y}|w\right)$ in local window $ w $,we propose a robust loss function to optimize the unsupervised network. The overall loss function is defined as

\begin{equation}
Loss\left(x_{1},x_{2},\hat{y}\right)=1-\frac{1}{\left|N\right|}\sum_{w=1}^{N}{\rm Scope}\left(x_{1},x_{2},\hat{y}|w\right),
\end{equation}
where, \emph{N} represents the total number of sliding windows in an image. The computed loss is back-propagated to train the network. The better performance of $\rm SSIM_{Y}$ is attributed to its objective function that maximizes structural consistency between the fused image and each of the input images.\par

\subsection{Implementation Details}
All the convolutional layers have 64 filters of size $ 3\times 3 $ in our proposed \textit{MFNet}. We randomly initialize the parameters of convolutional filters and pad zeros around the boundaries before applying convolution to keep the size of all feature maps the same as the input images. We use leaky rectified linear units (LReLUs)~\cite{maas2013rectifier} with a negative slope of 0.2 as the non-linear activation function except for the last convolutional (reconstruction) layer where we choose sigmoid as the activation function. For the feature extraction and reconstruction sub-networks, the number of  convolutional layers D1, D2 and D3 are set as 5, 6 and 7 respectively.\par

We use 60 pairs of multi-focus images from the benchmark Lytro Multi-focus Image dataset~\cite{nejati2015multi} and gray-scale Image dataset as our training data. Since the dataset is too small, we randomly crop $ 64\times 64 $ patches to form our final training dataset. The total number of the cropped patch is $50,000$. An epoch has 400 iterations of back-propagation. We use Tensorflow~\cite{abadi2016tensorflow} to train our model. %use Adam solver ~\cite{kingma2014adam}. 
In addition, we set the weight decay to $10e-4$, initialize the learning rate to $10e-3$ for all layers, set the decay coefficient to $ 10^{3}$ and the decay rate to $0.96$.

\setlength{\tabcolsep}{4pt}
\begin{table}[t]
	\begin{center}
		\caption{The objective assessment of different methods for the fusion of ``Clock'' source images.}
		\label{table:1}
		\begin{tabular}{ccccc}
			\hline\noalign{\smallskip}
			Methods & $ Q_{S} $ & $Q_{CV}$ & VIFF & EN\\ 
			\noalign{\smallskip}
			\hline
			\noalign{\smallskip}
			NSCT  & {\bf 0.9491} & 63.7236 & 0.9566 & 7.3278\\
			GF & 0.9444 & 75.0824 & 0.9319 & 7.2985\\
			DSIFT & 0.9447 & 71.5299 & 0.9410 & 7.3045\\
			BF & 0.9442 & 75.0824 & 0.9319 &7.2985 \\
			CNN & 0.9459 & 68.0495 & 0.7420  & 7.3077 \\
			\textit{MFNet} & 0.9362 & {\bf 98.3789} & {\bf 1.0588} & {\bf 7.5030} \\
			\hline
		\end{tabular}
	\end{center}
\end{table}
\setlength{\tabcolsep}{1.4pt}

\section{Experimental Results}
In this section, we compare the proposed \textit{MFNet} with several state-of-the-art multi-focus image fusion methods on benchmark datasets. We present quantitative evaluation and qualitative comparison. 

We compare the proposed method with five state-of-the-art multi-focus image fusion algorithms, including methods based on non-subsampled contournet transform (NSCT)~\cite{zhang2009multifocus}, guided filtering (GF)~\cite{li-kang2013image}, dense SIFT (DSIFT)~\cite{liu2015multi}, boundary finding (BF)~\cite{zhang2017boundary}, convolutional neural network (CNN)~\cite{liu2017multi}. We implemented these algorithms using codes acquired from their respective authors. We carry out extensive experiments on $40$ pairs of multi-focus images from two public benchmark datasets: 20 pairs from the multi-focus image fusion dataset~\cite{Liuyucode:2016}  and the other 20 pairs from a recently available dataset "Lytro"~\cite{Multi-focus:2016}.\par

Quantitative evaluation of  image fusion is not an easy task since it is often impossible to obtain the reference image. Thus, many evaluation metrics are introduced for evaluating image fusion performance. There is no consensus on which metrics can completely describe the fusion performance. We evaluate the multi-focus image fusion results using image structural similarity $Q_{S}$~\cite{piella2003new}, human perception $Q_{CV}$~\cite{chen2007human}, information entropy (EN)~\cite{kumar2015image} and visual information fidelity VIFF~\cite{han2013new}. Among these four evaluation metrics, the $Q_{S}$ and $Q_{CV}$ and VIFF are calculated from the input image pair and the resultant fused image, while the EN is calculated from fused image only.  $Q_{S}$ measures how well the structural information of the source images is preserved, $Q_{CV}$ measures how well the human perceive the results, VIFF measures the  visual information fidelity while EN estimates the amount of information present in the  fused image. For each of these metrics, the largest value indicates the best fusion performance.\par

\setlength{\tabcolsep}{4pt}
\begin{table}[t]
	\begin{center}
		\caption{The objective assessment of different methods for the fusion of ``Fence'' source images.}
		\label{table:2}
		\begin{tabular}{ccccc}
			\hline\noalign{\smallskip}
			Methods & $ Q_{S} $ & $Q_{CV}$ & VIFF & EN\\ 
			\noalign{\smallskip}
			\hline
			\noalign{\smallskip}
			NSCT  &  0.9314 & 27.7074 & {\bf 0.9316 } & 7.8515\\
			GF & 0.9273 & 24.1802 & 0.9175 & 7.8034\\
			DSIFT & 0.9210 & 28.6299 & 0.9283 & 7.8531\\
			BF & 0.9144 & { \bf 71.2070 } & 0.7897 & 7.8034 \\
			CNN & 0.9271 & 24.8961 & 0.9304  & 7.8034 \\
			\textit{MFNet} & {\bf 0.9385} &  34.7656 &  0.9294 & {\bf 7.8481} \\
			\hline
		\end{tabular}
	\end{center}
\end{table}
\setlength{\tabcolsep}{1.4pt}

\subsection{Comparison with other methods}

\begin{figure*}[t]
	\begin{center}
		\includegraphics[width=1.0\linewidth]{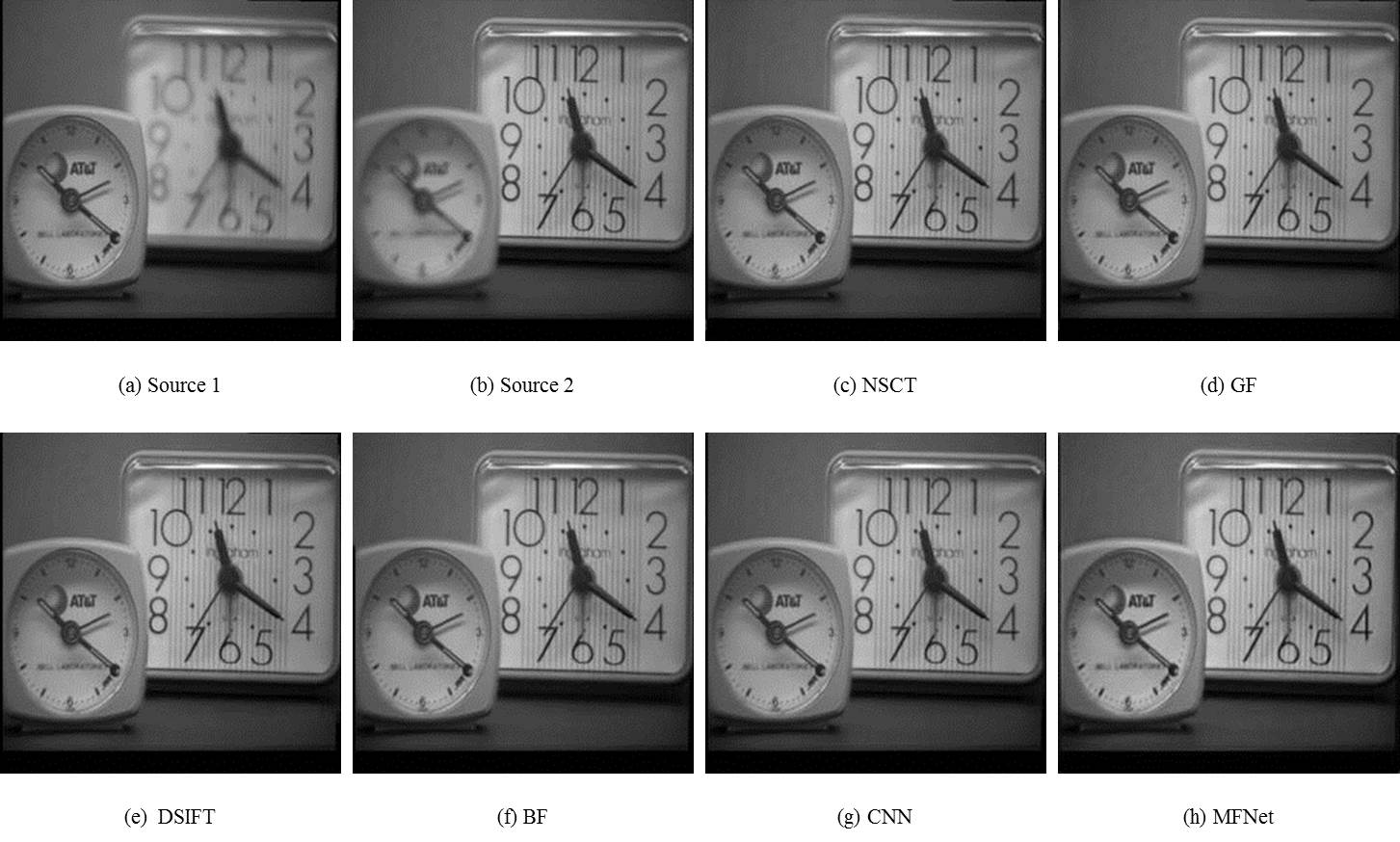}
	\end{center}
	\caption{ The ``Clock'' source image pair and their fused images obtained with different fusion methods.}
	\label{fig:4}
\end{figure*}

\begin{figure*}[h!]
	\begin{center}
		\includegraphics[width=1.0\linewidth]{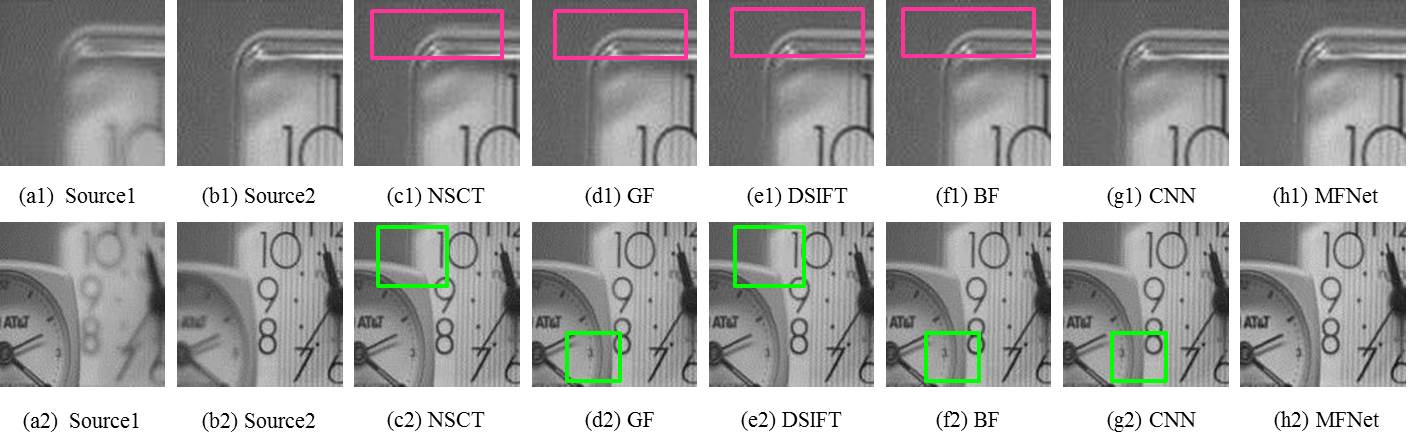}
	\end{center}
	\caption{ Magnified regions of the ``Clock'' source images and fused images obtained with different methods.}
	\label{fig:5}
\end{figure*}

Figure~\ref{fig:4} compares the results of our proposed \textit{MFNet} with other best performing multi-focus image fusion approaches on ``Clock'' image set. We can see that our proposed algorithm provides the best fusion result among these methods. For a better comparison, in Figure~\ref{fig:5} we depict the magnified regions of the fused images taken from Figure~\ref{fig:4}. The results clearly show that the fused images from \textit{MFNet} contain no obvious artifact in these regions, while the fused results from other methods contain some artifacts around the boundary of focused and defocused clocks (highlighted with green rectangles) and  pseudo-edges (highlighted with pink rectangles).\par

\begin{figure*}[t!]
	\begin{center}
		\includegraphics[width=1.0\linewidth]{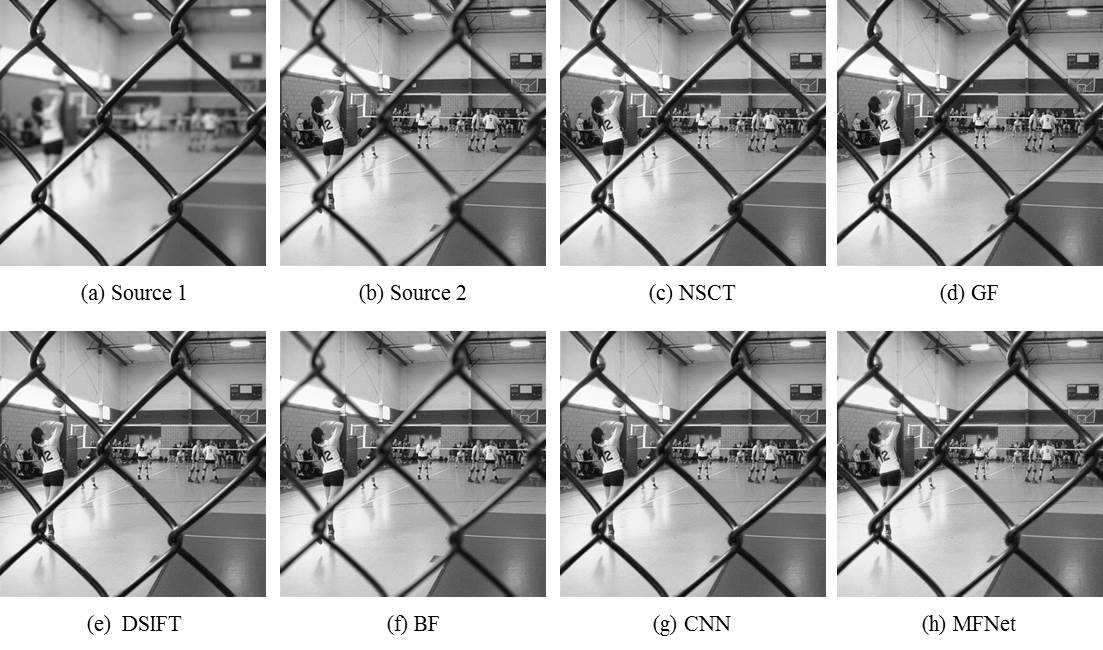}
	\end{center}
	\caption{ The ``Fence'' source image pair and their fused images obtained with different fusion methods.}
	\label{fig:6}
\end{figure*}

\begin{figure*}[h!]
	\begin{center}
		\includegraphics[width=1.0\linewidth]{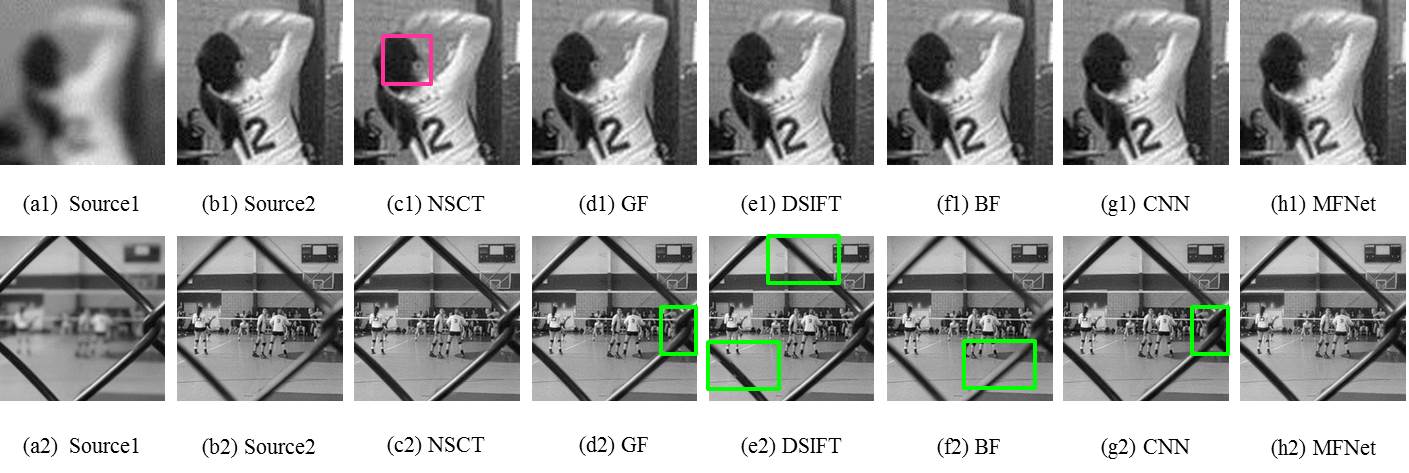}
	\end{center}
	\caption{  Magnified regions of the ``Fence'' source images and fused images obtained with different methods.}
	\label{fig:7}
\end{figure*}

In the second experiment, detailed results of ``Fence'' image set are shown in Figure~\ref{fig:6}. The fused result obtained with BF method is distinctly blurred. Once again magnified regions of these results are depicted in Figure~\ref{fig:7} for ease of comparison. Note that the fused result from the NSCT method contains artifacts (highlighted as pink rectangles) while the results of GF, DSIFT, BF and CNN algorithms suffer from blur artifact around the fence edges (Highlighted as green rectangles). However, the result obtained by our proposed algorithm are free from such artifacts.\par

\begin{figure*}[h!]
	\begin{center}
		\includegraphics[width=1.0\linewidth]{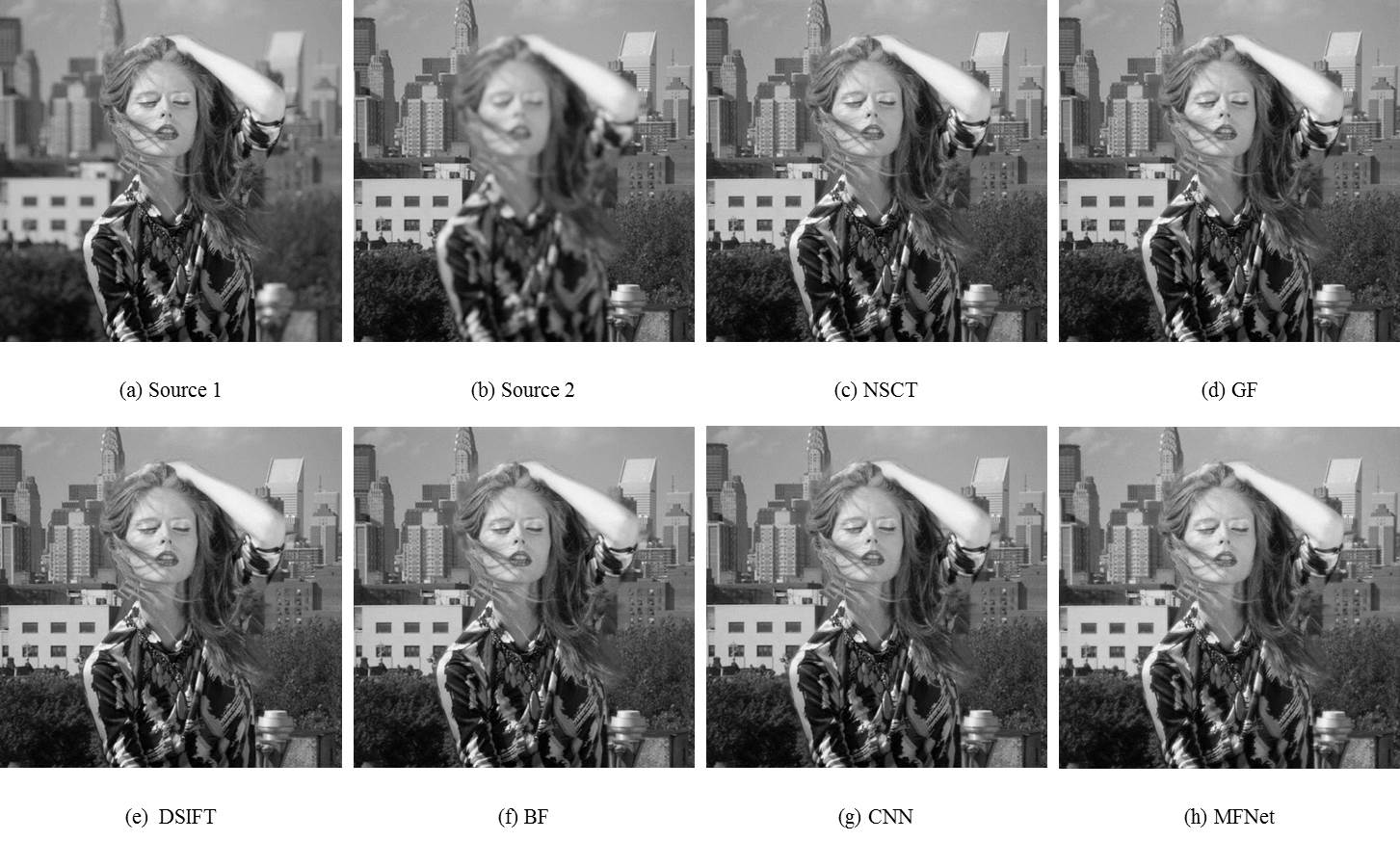}
	\end{center}
	\caption{ The ``Model Girl'' source image pair and their fused images obtained with different fusion methods.}
	\label{fig:8}
\end{figure*}

\begin{figure*}[h!]
	\begin{center}
		\includegraphics[width=1.0\linewidth]{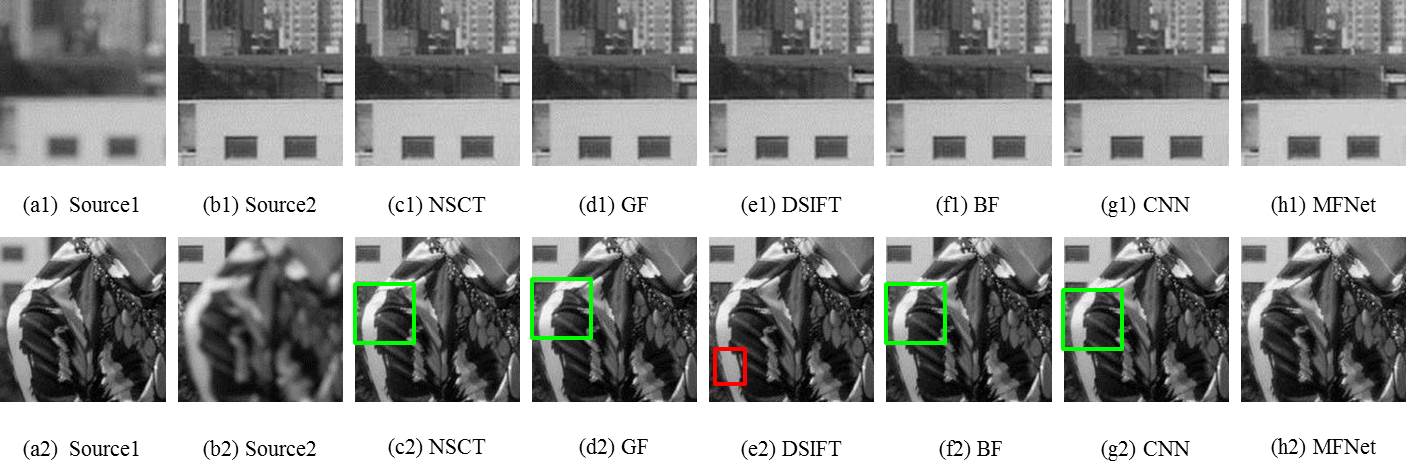}
	\end{center}
	\caption{ Magnified regions of the ``Model Girl'' source images and fused images obtained with different methods.}
	\label{fig:9}
\end{figure*}

Figure~\ref{fig:8} and Figure~\ref{fig:9} presents the original and magnified visual comparison of image fusion algorithms on ``Model Girl'' image set. Although all the algorithms show similar results for the background focused region (first row of Figure~\ref{fig:9}), we can clearly find blur artifacts in the girl's shoulder in the results of NSCT, GF, DSIFT, BF and CNN algorithms. The fused results from our method look more aesthetically pleasing.\par

The objective assessments of different methods for the fusion of the ``Clock'', ``Fence'' and ``Model Girl'' image sets are listed in Table~\ref{table:1},Table~\ref{table:2} and Table~\ref{table:3}, respectively, where the highest values are shown in bold. The results show that our proposed \textit{MFNet} outperforms the state-of-the-art in most cases using the four metrics. In some cases our proposed algorithm shows the second best performance. In general, only one metric can not objectively reflect the fused  quality, thus we use these four metric to objectively evaluate different methods. \par

\setlength{\tabcolsep}{4pt}
\begin{table}[t]
	\begin{center}
		\caption{The objective assessment of different methods for the fusion of ``Model Girl'' source images.}
		\label{table:3}
		\begin{tabular}{ccccc}
			\hline\noalign{\smallskip}
			Methods & $ Q_{S} $ & $Q_{CV}$ & VIFF & EN\\ 
			\noalign{\smallskip}
			\hline
			\noalign{\smallskip}
			NSCT  &  0.9357 & 15.1591 & 0.9612 & 7.7133\\
			GF & 0.9330 & 13.6080 & 0.9564 & 7.7133\\
			DSIFT & 0.9316 & 13.7120 & 0.9571 & 7.7110\\
			BF & 0.9298 & 13.9148 & 0.9523 & 7.7102 \\
			CNN & 0.9329 & 13.6979 & 0.9542 & 7.7099 \\
			\textit{MFNet} & {\bf 0.9371} &  {\bf 24.8138} &  {\bf 1.0011} & {\bf 7.7364} \\
			\hline
		\end{tabular}
	\end{center}
\end{table}
\setlength{\tabcolsep}{1.4pt}

\setlength{\tabcolsep}{5pt}
\begin{table}[h]
	\begin{center}
		\caption{The objective assessment of different methods for the fusion of ten pairs of  validation multi-focus source images.}
		\label{table:4}
		\begin{tabular}{cccccc}
			\hline\noalign{\smallskip}
			Dataset &Methods & $ Q_{S} $ & $Q_{CV}$ & VIFF & EN\\ 
			\noalign{\smallskip}
			\hline
			\noalign{\smallskip}
			\multirow{6}{*}{Data1} & NSCT  &  {\bf 0.9291} & 69.9239 & 0.9200 & 7.3454\\
			& GF & 0.9241 & 73.1916 & 0.8819 & 7.3350\\
			& DSIFT & 0.9218 & 76.5037 & 0.8776 & 7.3330\\
			& BF & 0.9222 & 77.1837 & 0.8740 & 7.3320 \\
			& CNN & 0.9234 & 76.6635 & 0.8783 & 7.3299 \\
			& \textit{MFNet} & 0.9201 &  {\bf 87.3684} &  {\bf 0.9771} & {\bf 7.4259} \\
			\multirow{6}{*}{Data2} & NSCT  & {\bf 0.9588}  & 11.0787 & 0.9652 & 7.4332\\
			& GF & 0.9578 & 6.2571 & 0.9574 & 7.4371\\
			& DSIFT & 0.9572 & 6.2545 & 0.9583 & 7.4377\\
			& BF & 0.9567 & 8.7372 & 0.9528 & 7.4358 \\
			& CNN & 0.9575 & 6.3135 & 0.9570 & 7.4369 \\
			& \textit{MFNet} &  0.9502 &  {\bf 25.4599} &  {\bf 1.0112} & {\bf 7.4869} \\
			\hline
		\end{tabular}
	\end{center}
\end{table}

%-------------

\begin{figure*}
	\begin{center}
		\includegraphics[width=1.0\linewidth]{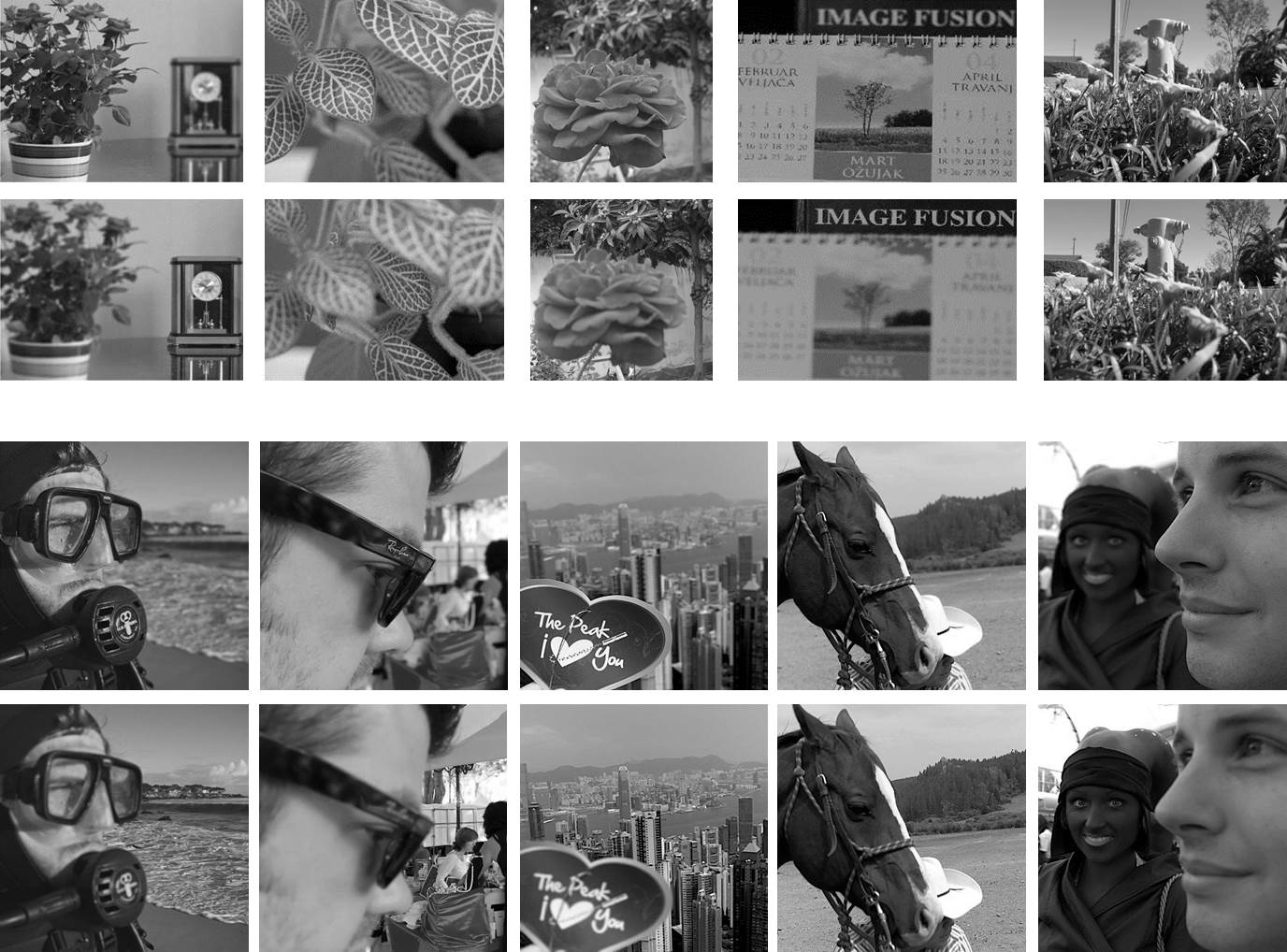}
	\end{center}
	\caption{ Ten pairs of multi-focus images used for validation}
	\label{fig:10}
\end{figure*}

\begin{figure*}
	\begin{center}
		\includegraphics[width=0.8\linewidth]{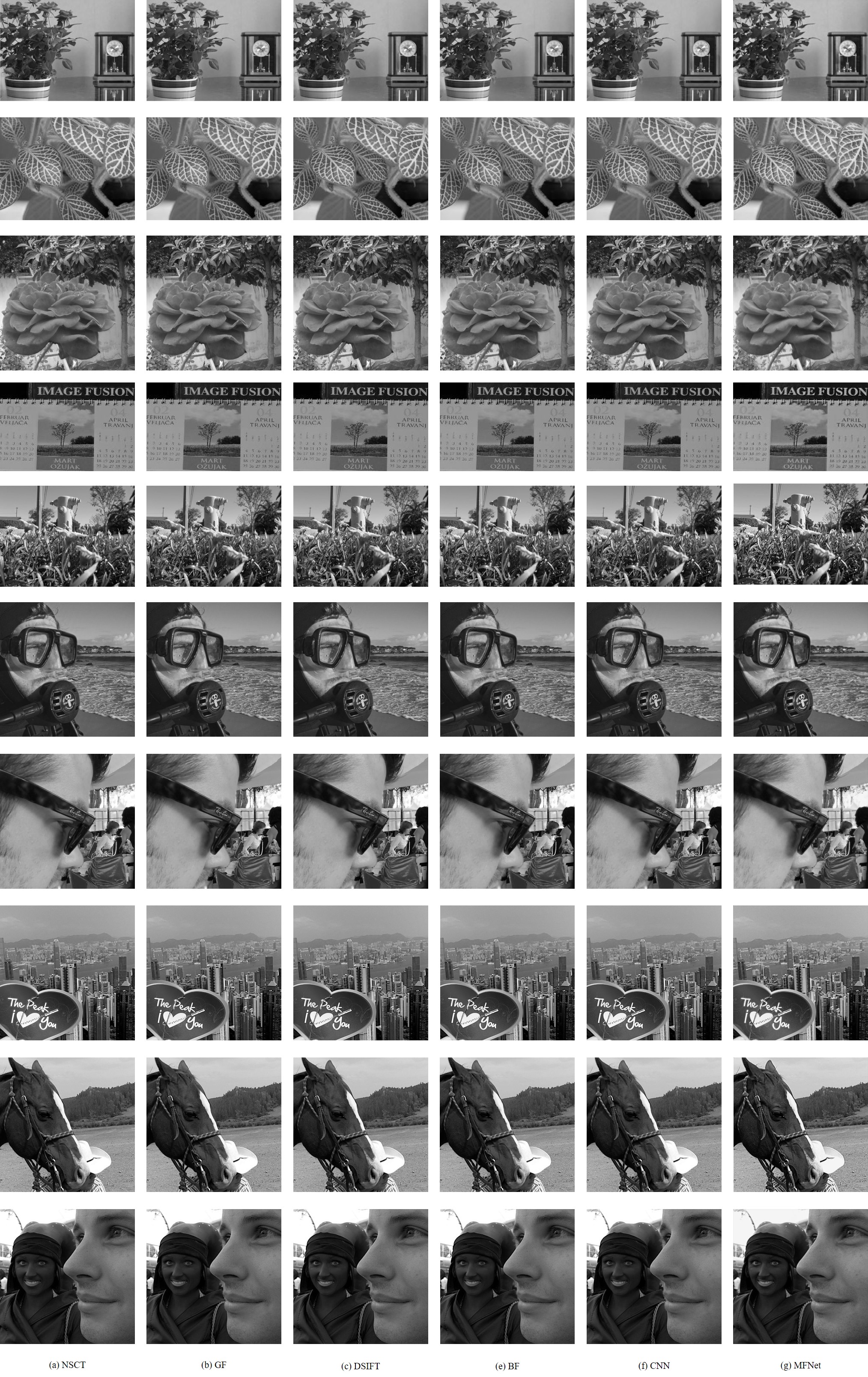}
	\end{center}
	\caption{ Fused results of ten pairs of source images obtained by different fusion methods.}
	\label{fig:11}
\end{figure*}

To further demonstrate the effectiveness of our proposed fusion method, ten pairs of popular multi-focus image sets are used, as shown in Figure~\ref{fig:10}. Among them five pairs are grayscale from~\cite{liu2017multi} (see in the first two rows of Figure~\ref{fig:10}) while the remaining from Lytro dataset. For convenience, we denote the first five pairs as Data1 and remaining as Data2. Figure~\ref{fig:11} depicts the results of different methods on the ten pair image set. Visual comparison of \textit{MFNet} with other image fusion methods shows that our proposed algorithm generates better quality fused images. The average scores achieved by the proposed and the compared fusion methods are reported in Table~\ref{table:4}. Our proposed method outperforms state-of-the-art fusion methods on all metrics except the $ Q_{S} $ metric.

\subsection{Execution time}
We use a desktop machine with 3.4GHz Intel i7 CPU (32 RAM) and NVIDIA Titan Xp GPU (12 GB Memory) to evaluate our algorithm.  We choose multi-focus image pairs with a spatial resolution of $256\times 256$ ,$ 320\times 240$ and $520\times 520$ respectively and evaluate our method as well as the CNN based method~\cite{liu2017multi} using these three pairs of images. The average runtime of our proposed \textit{MFNet} for $256\times 256$ ,$ 320\times 240$ and $520\times 520$ size images is 3.7s, 4.1s and 5.1s respectively. This runtime is significantly lower than that of ~\cite{liu2017multi} which takes 54.8s, 46.62s and 115.8s respectively to fuse the same size images.

\section{Conclusion}

We introduced an end-to-end approach for multi-focus image fusion that learns to directly predict the fusion image from an input pair of images with varied focus. Our model directly predicts the fusion image using a  deep unsupervised network (\textit{MFNet}) which employs the structural similarity (SSIM) image quality metric as a loss function. To the best of our knowledge, \textit{MFNet} is the first ever unsupervised end-to-end deep learning method to perform multi-focus image fusion. The proposed model extracts a set of common low-level features from each input image. Feature pairs of the input images are fused and combined with features extracted from the average of the input images to generate the final representation or feature map. Finally, this  representation is passed through a feature reconstruction network to get the final fused image. We train our model on a large set of images from multi-focus image sets and  perform extensive quantitative and qualitative evaluations to demonstrate the efficacy of our proposed algorithm.

%I wish you the best of success.

\ifCLASSOPTIONcompsoc
\section*{Acknowledgments}
\else
\section*{Acknowledgment}
\fi

This research was supported by the China Scholarship Council (CSC), Natural Science Foundation of Shaanxi Province(2017JM6079), Joint Foundation of the Ministry of Education of the People’s Republic of China (614A05033306) and ARC Discovery Grant DP160101458. We thank NVIDIA for the GPU donation. 

\ifCLASSOPTIONcaptionsoff
\newpage
\fi

%\end{thebibliography}

%\bibliographystyle{IEEEtran}
%\bibliography{Deepfusion}

\begin{IEEEbiography}[{\includegraphics[width=1in,height=1.25in,clip,keepaspectratio]{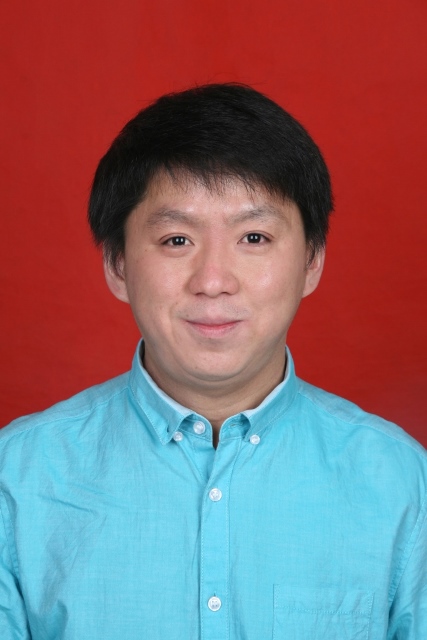}}]{Xiang Yan}(S'17) received his M.S. degree of Electronics Science and Technology from Xidian University. He is currently pursuing doctoral degree in Physical Electronics, Xidian University. He has been a visiting Ph.D. student at the School of Computer Science and Software Engineering, the University of Western Australia. His current research is focused on image fusion, action detection and recognition, deep learning.
\end{IEEEbiography}


\begin{IEEEbiography}[{\includegraphics[width=1in,height=1.25in,clip,keepaspectratio]{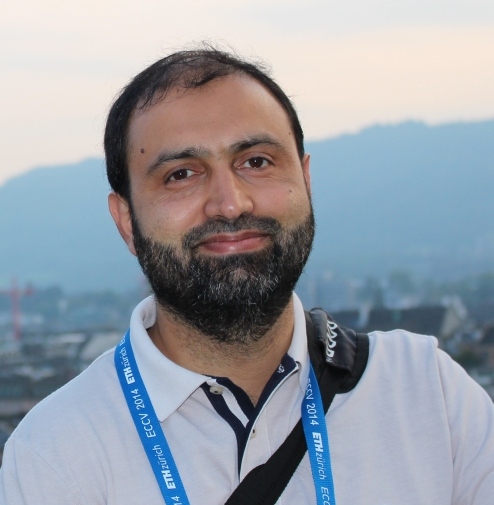}}]{Syed Zulqarnain Gilani}
received his Phd in 3D facial analysis from the University of Western Australia. Earlier, he completed his B.Sc Engineering from National University of Sciences and Technology (NUST), Pakistan and MS in Electrical Engineering from the same university in 2009, securing the Presidents Gold Medal. He served as an Assistant Professor in NUST before joining UWA. He is currently a post-doc Research Associate in Computer Science and Software Engineering at UWA. His research interests include 3D Morphometric Face Analysis, pattern recognition, machine learning and video description.
\end{IEEEbiography}

\begin{IEEEbiography}[{\includegraphics[width=1in,height=1.25in,clip,keepaspectratio]{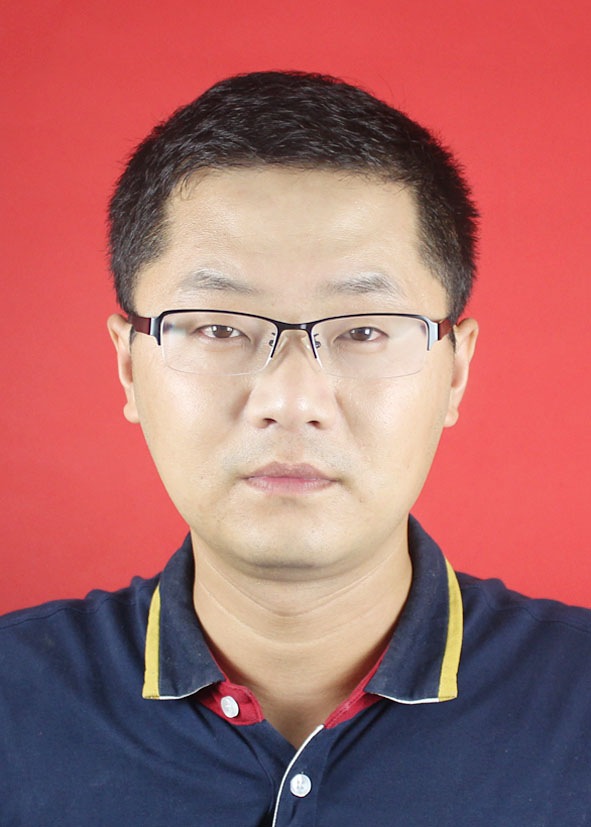}}]{Hanlin Qin}
received the B.Eng. and Ph.D. degree from Xidian University, Xi’an, China, in 2004 and 2010 respectively. He has authored over 60 papers. He received a Technology Invention Award from the Ministry of Education of the People’s Republic of China in 2106. He is currently an Associate Professor of School of Physics and Optoelectronic Engineering, Xidian University. His current research interests include photoelectric imaging, image processing, visual tracking, target detection and recognition and deep learning.
\end{IEEEbiography}

%\newpage

\begin{IEEEbiography}[{\includegraphics[width=1in,height=1.25in,clip,keepaspectratio]{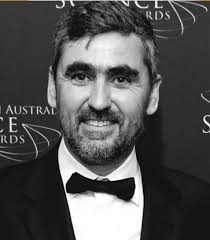}}]{Ajmal Mian}
is an Associate Professor of Computer Science at The  University  of  Western  Australia. He completed  his  PhD  from the same institution in  2006  with  distinction  and  received  the  Australasian  Distinguished Doctoral  Dissertation  Award  from  Computing  Research  and  Education  Association  of  Australasia.
He  received  the  prestigious  Australian  Postdoctoral and  Australian  Research  Fellowships  in  2008  and 2011  respectively.  He  received  the  UWA  Outstanding  Young  Investigator  Award  in  2011,  the  West Australian Early Career Scientist of the Year award in  2012, the  Vice-Chancellors  Mid-Career  Research  Award  in  2014 and the Aspire Professional Development Award in 2016.  He  has  secured  seven  Australian  Research  Council grants, a National Health and Medical Research Council grant and a DAAD German Australian Cooperation grant. He has published over 150 scientific papers.  His  research interests  include  computer  vision,  machine  learning,  face  recognition,  3D shape analysis and hyperspectral image analysis
\end{IEEEbiography}


%\vfill



\end{document}